\documentclass{article}

\usepackage[preprint]{neurips_2024}

\usepackage[utf8]{inputenc} % allow utf-8 input
\usepackage[T1]{fontenc}    % use 8-bit T1 fonts
\usepackage{hyperref}       % hyperlinks
\usepackage{url}            % simple URL typesetting
\usepackage{booktabs}       % professional-quality tables
\usepackage{amsfonts}       % blackboard math symbols
\usepackage{nicefrac}       % compact symbols for 1/2, etc.
\usepackage{microtype}      % microtypography

\usepackage[accsupp]{axessibility}  % Improves PDF readability for those with disabilities.

\usepackage{graphicx}
\usepackage{subfig}
\usepackage{bm}
\usepackage{ulem} 
\usepackage{float}
\usepackage{geometry}

\usepackage{hyperref}

\usepackage{orcidlink}

\usepackage{amstext}

\usepackage{amssymb}

\usepackage{bm}
\usepackage{cases}
\usepackage{color}
\usepackage{ulem}

\usepackage{url}
\usepackage{epsfig}
\usepackage{amsthm}
\usepackage{multirow}
\usepackage{soul}
\usepackage{footnote}
\usepackage{tablefootnote}
\usepackage{caption}
\usepackage{colortbl}

\usepackage{pifont}
\usepackage{algorithm}
\usepackage{mathtools}
\usepackage{algpseudocode}
\usepackage{listings}
\usepackage{wrapfig}
\usepackage{subcaption}
\usepackage{xcolor,colortbl}

\definecolor{top1}{RGB}{255,179,179}
\definecolor{top2}{RGB}{255,217,179}
\definecolor{top3}{RGB}{255,255,179}

\usepackage{tikz}
\usepackage{comment}
\usepackage{amsmath} % define this before the line numbering.}
\usepackage{bbm}

\usepackage{bm}

\newcommand*\colourcheck[1]{%
  \expandafter\newcommand\csname #1check\endcsname{\textcolor{#1}{\ding{52}}}%
}
\colourcheck{blue}
\colourcheck{green}
\colourcheck{red}
\usepackage{makecell}

\newcommand{\model}{\textit{{GaussianStego}}}
\newcommand{\Ste}{{{Steganography}}}

\newcommand{\obj}{{{Objaverse}}}

\title{
GaussianStego: A Generalizable Stenography Pipeline for Generative 3D Gaussians Splatting
}

\author{
  {Chenxin Li\textsuperscript{1}\thanks{Equal contribution}, 
  Hengyu Liu\textsuperscript{1$\ast$},
Zhiwen Fan\textsuperscript{2}, 
Wuyang Li\textsuperscript{1}, 
Yifan Liu\textsuperscript{1},
Panwang Pan\textsuperscript{3},
Yixuan Yuan\textsuperscript{1}}\\
  {\textsuperscript{1}The Chinese University of Hong Kong}, \, {\textsuperscript{2}University of Texas at Austin}, \, {\textsuperscript{3}ByteDance}\\
}

\begin{document}

\maketitle

\begin{abstract}
Recent advancements in large generative models and real-time neural rendering using point-based techniques pave the way for a future of widespread visual data distribution through sharing synthesized 3D assets. However, while standardized methods for embedding proprietary or copyright information, either overtly or subtly, exist for conventional visual content such as images and videos, this issue remains unexplored for emerging generative 3D formats like Gaussian Splatting. 
We present \model, a method for embedding steganographic information in the rendering of generated 3D assets. Our approach employs an optimization framework that enables the accurate extraction of hidden information from images rendered using Gaussian assets derived from large models, while maintaining their original visual quality. We conduct preliminary evaluations of our method across several potential deployment scenarios and discuss issues identified through analysis. \model~represents an initial exploration into the novel challenge of embedding customizable, imperceptible, and recoverable information within the renders produced by current 3D generative models, while ensuring minimal impact on the rendered content's quality.
Project website:~\url{https://gaussian-stego.github.io/}.
\end{abstract}

\section{Introduction}

% The capabilities of automatic 3D content generation extend across diverse domains, including digital gaming, virtual reality, and cinematic production~\cite{samavati2023deep, tang2023dreamgaussian}. Foundational techniques, such as image-to-3D and text-to-3D, as highlighted in studies like \cite{poole2022dreamfusion,tang2024lgm,liu2024one}, offer substantial benefits by significantly reducing the manual labor traditionally required from professional 3D artists. These innovative approaches simplify the creation process, making it more accessible and time-efficient. This advancement fosters a more inclusive environment in the field of 3D design and modeling. It democratizes the creation process, enabling individuals without expertise or professional training in 3D artistry to actively engage in and contribute to the production of 3D assets.
Automatic 3D content generation has revolutionized diverse fields including gaming, virtual reality, and film production \cite{samavati2023deep, tang2023dreamgaussian}. Foundational techniques like image-to-3D and text-to-3D \cite{poole2022dreamfusion,tang2024lgm,liu2024one} significantly reduce the manual labor required from professional 3D artists. These approaches simplify and democratize the creation process, enabling individuals without specialized expertise to contribute to 3D asset production. By making 3D design more accessible and efficient, these innovations foster a more inclusive environment in the field, potentially reshaping the landscape of 3D content creation and distribution. Furthermore, this democratization opens up new possibilities for creative expression and innovation, as a wider range of perspectives and ideas can now be translated into 3D form, potentially leading to more diverse and rich content across various media platforms.

Research on 3D generation has evolved from Score Distillation Sampling (SDS) methods \cite{poole2022dreamfusion, lin2023magic3d, liu2023zero, tang2023dreamgaussian, chen2024survey}, which create detailed 3D objects from text or single-view images by optimizing 3D representations to match pretrained 2D diffusion model predictions. While SDS techniques produce impressive results, they often face challenges related to generation speed and diversity. More recent advancements have led to efficient feed-forward 3D-native techniques, trained on large-scale 3D datasets \cite{deitke2023objaverse,deitke2023objaversexl}, capable of generating 3D assets within seconds. The latest research, employing Gaussian Splatting and optimized 3D backbones \cite{tang2024lgm}, further improves texture detail and geometric complexity. As we witness this rapid progress in 3D asset generation, a new question emerges: \textit{Can we apply steganography to invisibly mark the upcoming profusion of generated 3D assets?}

% The established digital steganography method~\cite{cox2007digital} primarily concentrates on embedding concealed messages within 2D images. The recent surge in generative AI and social media platforms has led to both an immense proliferation of generated images and videos shared online and practical requirement of steganography for mainstream generative content such as 2D images~\cite{tancik2020stegastamp} or videos~\cite{sadek2015video}, in an effort to prevent content misuse. A series of works are either dedicated to enabling users and data providers to protect their copyright, embedding implicit information such as ownership and source into generated content for traceability detection~\cite{baluja2017hiding}, or committed to preventing the misuse of content by embedding covert backdoors into genuine content~\cite{jing2021hinet}, thereby preventing the contents being re-created and misused by generative models.

Traditional digital steganography methods \cite{cox2007digital} primarily focus on embedding hidden messages within 2D images. The recent surge in generative AI and social media platforms has led to an explosion of generated content shared online, driving practical applications of steganography for mainstream generative media. This trend has spurred research into enabling users and data providers to protect their intellectual property by embedding ownership information and source data into generated content for traceability. Additionally, efforts have been made to prevent content misuse through the embedding of covert backdoors, rendering the content unusable for unauthorized re-creation by generative models \cite{baluja2017hiding, baluja2019hiding}. These advancements in steganography aim to address the growing concerns surrounding copyright protection and content abuse in the era of widespread AI-generated media.

With ongoing advances in 3D representations powered by large generative models, we anticipate a future where sharing captured 3D content online becomes as commonplace as sharing 2D images and videos today. This emerging trend prompts us to explore the following research questions:
\ding{202} Injecting information into 2D generated contents like images and videos for copyright or ownership identification is a well-established pipeline~\cite{tancik2020stegastamp,sadek2015video}, but can we develop pipelines to preserve such information when people share and render 3D scenes through generated 3D Gaussians?
\ding{203} Existing 3D watermarking techniques for 3D representations like NeRF and mesh tend to fail on unseen objects~\cite{li2023steganerf,luo2023copyrnerf,jang2024waterf}, but can we generalize the ability to watermarking the representation even of unseen objects resorting to the general prior knowledge in foundation models?
\ding{204} Common image steganography methods~\cite{zhu2018hidden,tancik2020stegastamp} designed for image generation (e.g. Stale Diffusion) typically embed a message string, hidden image, or covert backdoor into an image, but can we embed a wider range of signal modalities into generated Gaussians?

% Now, with the continuous advancements in foundation models for 3D generation, 3D assets are becoming inexpensive and widespread. We anticipate a future where people will increasingly share their captured 3D content online, akin to the current sharing of 2D images and videos. Consequently, we are eager to investigate the following research questions: 
% \ding{202} Injecting information into 2D generated content such as images and videos for copyright or ownership identification is a common practice with established pipelines~\cite{tancik2020stegastamp,sadek2015video}. However, can we design feasible pipelines to preserve this information when people generate, share and render 3D scenes through generative gaussians?
% \ding{203} Current 3D watermarking methods for representations like NeRF and mesh are often not generalizable~\cite{li2023steganerf,luo2023copyrnerf,jang2024waterf}. However, can we leverage the generalization capabilities of 3D generative base models to enable a single training session for steganography to be applicable across unseen 3D objects?
% \ding{204} Common image steganography methods~\cite{zhu2018hidden,tancik2020stegastamp} designed for generative models embed a message string, hidden image, or covert backdoor into a what generated. But can we allow a broader range of signal modalities hidden into the generated Gaussians?

Driven by these inquiries, we propose a steganographic framework for a generalizable 3D Gaussian generative model. This framework seamlessly integrates customizable, imperceptible, and recoverable information into the generative process without compromising visual quality or the generation pipeline.
Unlike traditional image steganography that embeds hidden signals into specific source images, our approach aims to recover the intended hidden signal from generated 3D Gaussians rendered from predetermined viewpoints for verification purposes.
Specially, there is an embedding phase in \model~utilizing a vision foundation model to extract informative watermark embeddings, injecting them into the intermediate features of a 3D Gaussian generation baseline via cross-attention. For recovery, a U-Net-based decoder retrieves the concealed information from images rendered in specific verification viewpoint.
We employ adaptive gradient harmonization to constrain updates to weights where gradients of information hiding and rendering loss align, balancing rendering preservation and information hiding. Extensive experiments confirm our model's superiority in novel view synthesis rendering quality and high-fidelity transmission of hidden information.
Our contributions are summarized as follows: 
\begin{itemize}
\item{We elevate content generation steganography to a new level focused on 3D assets, pioneering the embedding of customizable, imperceptible, and recoverable information in generated Gaussian representations.}
\item{We propose an adaptive gradient harmonization strategy to guide the injected hidden information towards weights that exhibit greater harmony between rendering and hidden information gradients, balancing steganography and rendering quality objectives.}
\item{We empirically validate our framework on 3D objects across different domains with hiding multi-modal secrete signals,
, achieving high recovery accuracy without compromising rendering quality.}
\end{itemize}

\section{Related Work}

\textbf{Generative 3D Gaussian Splatting.}
Gaussian splatting, introduced by Kerbl et al.\cite{kerbl20233d}, is a notable 3D representation known for its expressiveness and rendering efficiency. While enhancing detail requires careful initialization and strategic densification\cite{chen2023gsgen,yi2023gaussiandreamer}, our research explores a feed-forward approach for autonomous generation of 3D Gaussians.
Unlike SDS-based optimization techniques, 3D-native feed-forward methods trained on large datasets can produce 3D models in seconds~\cite{deitke2023objaverse,deitke2023objaversexl}. Various studies have explored text-conditioned diffusion models for 3D formats~\cite{nichol2022point,jun2023shap,liu2023meshdiffusion,muller2023diffrf,chen2023primdiffusion,cao2023large,chen2023single,wang2023rodin,zhao2023michelangelo,yariv2023mosaic,liu2024endogaussian,li2024endora}, though these often face scaling challenges or yield lower quality outputs.
Recent advancements include regression models for rapid NeRF prediction from single-view images~\cite{hong2023lrm}, and Instant3D~\cite{li2023instant3d}, which combines text-to-multi-view image diffusion with multi-view LRM for fast, diverse text-to-3D generation. Further LRM extensions explore pose prediction~\cite{wang2023pflrm}, diffusion integration~\cite{xu2023dmv3d}, and human-centric datasets~\cite{weng2024single}.
As 3D Gaussian Splatting matures and enhances 3D asset accessibility, it's timely to explore steganography tailored for this emerging field.

\textbf{\Ste~in Image Diffusion Model.}
Deep learning advancements have significantly propelled deep watermarking. Works by \cite{hayes_generating_2017} and \cite{zhu_hidden_2018} introduced end-to-end learning paradigms where watermark encoders and decoders are refined via adversarial objectives, enhancing transmission fidelity and robustness. \cite{zeng2023securing} further extended this by concurrently optimizing a watermarked encoder and its detector using image datasets.
Recent methodologies, such as those by \cite{yu_artificial_2022}, implement a two-stage process, integrating watermark encoding within the generative framework. The Stable Signature technique \cite{fernandez_stable_2023} applies this concept to latent diffusion models, fine-tuning the latent decoder to work with a pre-trained watermark encoder. Similarly, \cite{zhao_recipe_2023} adopts this approach for unconditional diffusion models.
Distinct from traditional techniques, methods for language models like those proposed by \cite{kirchenbauer_watermark_2023} modify the output distribution to embed watermarks directly into generated data distributionally, without explicit training.
While these developments were crucial for traditional media formats, the rising prominence of point-based Gaussian representations for 3D scenes necessitates the extension of steganographic techniques from 2D to 3D geometric scenes. This transition is poised to become a critical area of research in the near future, reflecting the evolving landscape of visual data representation.

\textbf{\Ste~in 3D Representation.}
Within 3D watermarking, traditional approaches by Ohbuchi et al.\cite{ohbuchi2002frequency}, Praun et al.\cite{praun1999robust}, and Wu et al.~\cite{wu20153d} relied on Fourier or wavelet transformations applied to mesh structures. Recent innovations have expanded the field's scope:
Hou et al.\cite{hou2017blind} exploited 3D printing artifacts for watermarking, while Son et al.\cite{son2017perceptual} and Hamidi et al.\cite{hamidi2019blind} utilized mesh saliency to minimize vertex distortions and enhance robustness. Liu et al.\cite{liu2019novel} explored watermarking for point clouds, focusing on vertex curvatures.
A significant advancement came from Yoo et al.\cite{Yoo_2022_CVPR}, who introduced a deep-learning method to embed messages in 3D meshes and extract them from 2D renderings. StegaNeRF\cite{li2023steganerf} further pioneered embedding messages into neural radiance fields (NeRF), enabling extraction of multimodal information from 2D renderings.
While these techniques primarily address explicit 3D models or implicit representations like NeRF, emerging technologies such as point-based Gaussian splatting are gaining attention, signaling a potential shift in 3D watermarking research and applications.

\section{Preliminary}
% \subsection{Preliminaries}

\textbf{3D Gaussian Splatting.}
Gaussian splatting~\cite{kerbl20233d} employs a collection of 3D Gaussians to represent 3D data. 
Specifically, each Gaussian is defined by a center $\mathbf{x} \in \mathbb R^3$, a scaling factor $\mathbf{s} \in \mathbb R^3$, and a rotation quaternion $\mathbf{q} \in \mathbb R^4$. 
Additionally, an opacity value $\alpha \in \mathbb R$ and a color feature $\mathbf{c} \in \mathbb R^C$ are maintained for rendering, where spherical harmonics can be used to model view-dependent effects.
These parameters can be collectively denoted by ${\boldsymbol{\Omega}}$, with ${\boldsymbol{\Omega}}_i = \{\mathbf{x}_i, \mathbf{s}_i, \mathbf{q}_i, \alpha_i, \mathbf{c}_i\}$ representing the parameters for the $i$-th Gaussian. 
Rendering of the 3D Gaussians involves projecting them onto the image plane as 2D Gaussians and performing alpha composition for each pixel in front-to-back depth order, thereby determining the final color and alpha. 

\textbf{Generative Gaussian Splatting.}
% multi-view diffusion
Traditional 2D diffusion models generate images from a single viewpoint, lacking 3D viewpoint capabilities. Recent methods fine-tune multiview diffusion models on 3D datasets, incorporating camera pose as input. This enables generating multiview images from text or single-view images, generalizing to unseen objects. With efficient 3D representations like Gaussian Splatting emerging, researchers are extending generalizable multiview diffusion pipelines to include 3D spatial points as input during training. An additional consistency loss encourages view-consistent image generation, enhancing performance when applied in 3D Gaussians.

\section{Method}

\begin{figure*}[!t]   
	\centering	   \includegraphics[width=1.0\linewidth]{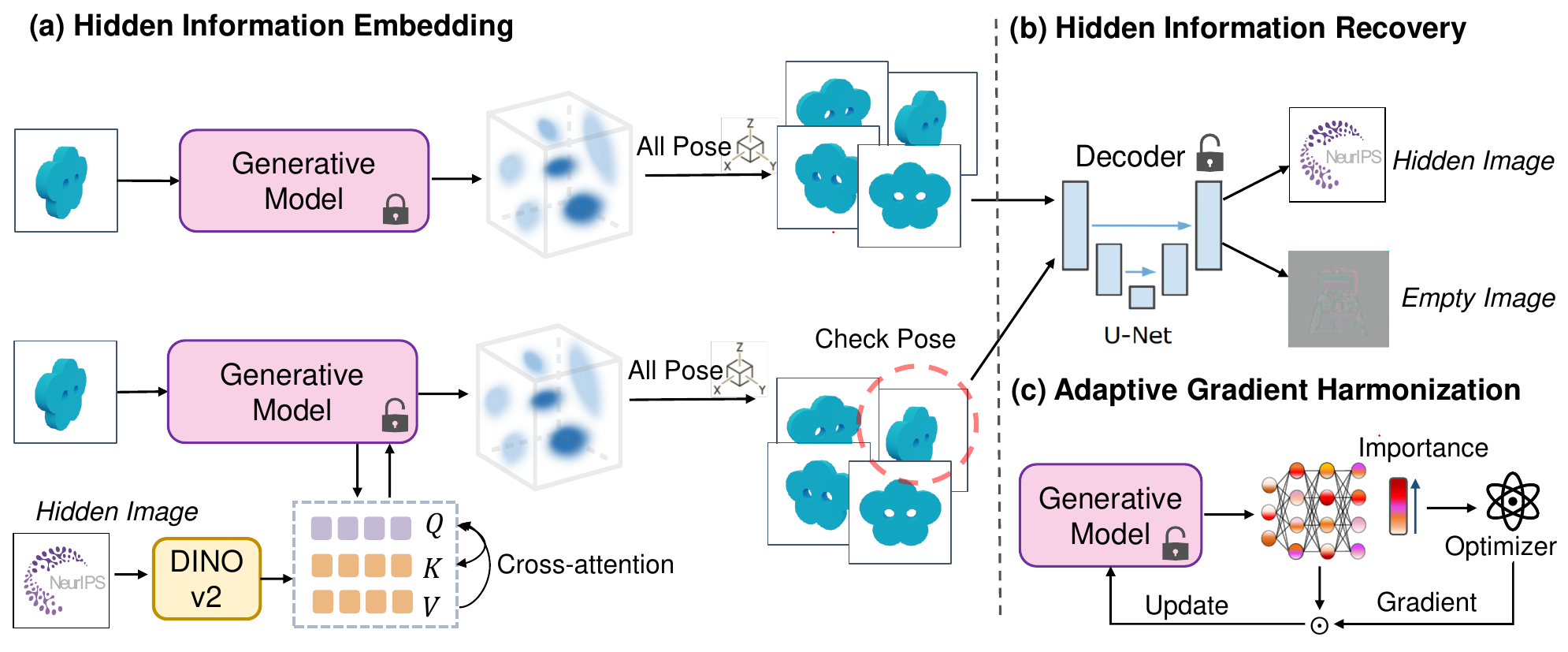}     
 	\vspace{0.75em}
	\caption{ 
 \model~training overview. 
{
During (a) Hidden Information Embedding, \model~incorporates the DINOv2 features of the hidden information into the intermediate feature of Gaussian generation via cross-attention. In (b) Hidden Information Recovery, a U-Net-based decoder is employed to retrieve the hidden information from the rendered image under the checking pose. Through the optimization process, (c) Adaptive Gradient Harmonization is utilized to maintain a balance between the rendering and hidden recovery.
% At the first stage (a) Hidden Information Embedding, \model~utilize cross-attention to get hidden information embedding to Gaussian. In the stage of recovery (b), a U-Net based detector is used to recover hidden information from rendering image under the Check Pose. During optimization, (c) Adaptive Gradient Harmonization is used to maintain the rendering quality of objects.
 % At the first stage, we optimize ${\boldsymbol{\Theta}_{0}}$ with standard NeRF training. At the second stage, we initialize ${\boldsymbol{\Theta}}$ with ${\boldsymbol{\Theta}_{0}}$ and optimize for the steganography objectives. We train the decoder ${F_{\psi}}$ to recover hidden information from \model~renderings and no hidden information from original NeRF renderings. We introduce {\it Classifier Guided Recovery} to improve the accuracy of recovered information, and {\it Adaptive Gradient Masking} to balance between steganography ability and rendering visual quality.
 }
 }
  \label{fig:ppline}    
\end{figure*}

% \subsection{Steganography of Generative Gaussian Splatting }
\textbf{Pipeline Overview.}
Conventional Image-to-3D generative models $F_{\boldsymbol{\Theta}}$, convert an input reference image $I$ into a generative 3D Gaussians, which can be parameterized as $\boldsymbol{\Omega} = F_{\boldsymbol{\Theta}}(I)$.
% , as the generated representation is only conditioned on the parameters of generative model and the reference image.
The objective of our \model~is to discreetly incorporate proprietary steganographic information into the Gaussian process generated by the generative model during the generation process, thereby eliciting virtually imperceptible visual changes in the rendering of the generated Gaussian.
Specifically, given the hidden image information $H$ that needs to be embedded, there is a hidden information embedding stage and a hidden decoding stage. During the embedding process, we employ an hidden embedder $F_{\boldsymbol{\Phi}}$  to inject and integrate the features of the hidden image $f_H$ into the intermediate feature $f_I$ of the generation process. 
In the decoding process, we specify a checking view $P_c$ among all available poses $P$.
% $\{P_i\}_{i=1}^N$. 
When rendering from the check view using the generated Gaussian representation $\boldsymbol{\Omega}$, our objective is to recover the embedded information $S$ from the rendering results via a decoder with learnable weights $\boldsymbol{\Psi}$. Fig.~\ref{fig:ppline} presents an overview of the proposed \model~framework.

\subsection{Hidden Information Embedding}
Models that are operated under the demand to predict new views from a single input are inherently ill-posed, given that atypical 2D images can be projected from entirely distinct 3D representations. Consequently, when there is a requirement to embed 2D images into 3D representations, the anticipated 3D generative model is optimized to ensure that 2D images carrying hidden information can be inversely projected into the underlying 3D space in conformity with human perceptual understanding. Under such circumstances, directly embedding 2D watermark images into the feed-forward process of the equally ill-posed 3D generative model could potentially have deleterious effects on this inverse projection process.
Inspired by the robust dense prediction capabilities exhibited by recent visual backbone models, particularly the semantic consistency demonstrated by spatial representations derived from the DINOv2 encoder compared to other encoders like CLIP~\cite{liu2023zero, liu2023syncdreamer, radford2021learning}, we advocate extrapolating the representational capabilities of the DINOv2 encoder to the to-be-encoded implicit watermark information. This effectively extracts the features of hidden information as $f_H$.

Although features derived from the visual base model provide a robust means of projecting the image into 3D space, this process can still result in the loss of detailed visual watermarks, thus impacting the visual quality of the hidden information. To address this challenge, we propose an early injection of the spatial details of the hidden information into the intermediate features derived from the image-to-3D Gaussian generation process. By integrating the image watermark into the intermediate feature $f_I$ of the generation process through cross-attention, the hidden image can effectively influence the update of the 3D representation via adaptive attention adjustment. Specially, the intermidate image feature $f_I$ can be updated by: $f_{I} :=\operatorname{Softmax}(\frac{K \cdot Q^T}{\sqrt{d}}) \cdot V$ (where $d$ denotes the length of the key and query features), with the key $K$, query $Q$ and value $V$ can be obtained by:
\begin{equation}
% \begin{split}
    % V^{new}_{s} = softmax(\frac{K_{s}Q_{c}^T}{\sqrt{d}}) \cdot V_{s}, 
    % \\ 
    % K = [K_I, K_0], V = [V_I, V_0],
% \end{split}
% f_{I} :=
% \operatorname{Softmax}(\frac{K(f_H) \cdot Q(f_{I})^T}{\sqrt{d}}) \cdot V(f_H),
K = F_{\boldsymbol{\Phi}_K}(f_H), \quad V= F_{\boldsymbol{\Phi}_V}(f_H), \quad Q = F_{\boldsymbol{\Phi}_Q}(f_I)
\end{equation}
% where the keys 'K' and values 'V' are derived from the image watermark, while the query 'Q' originates from the features of the 3D model by the input CHECK view. As a result, within the context of cross-attention injection, we leverage the features of the image watermark to weight the 3D features via cross-attention. This approach facilitates the propagation of visual cues from the image watermark through the network transit.
where $F_{\boldsymbol{\Phi}_*}$ respectively for key $K$ and value $V$ represent the linear transformation layer applied to the feature $f_H$ derived from the hidden images, while that for  query $Q$ represents the linear transformation layer applied to the feature $f_I$.
Therefore, in the context of cross attention injection, we utilize the features of image watermarks to weight 3D features through cross attention. Effectively, the strategy promotes the propagation of visual cues from image watermarks through network propagation~\cite{li2024u,li2022hierarchical,zhang2021generator}.

\subsection{
Hidden Information Recovery
}
% \paragraph{Contrastive Watermarking}
% 如何施加loss的，如何指定view的?
Given the set of camera poses $P$
% $\{P_i\}_{i=1}^N$ 
for the image rendered with the generated 3D Gaussians, we aim to extract the concealed information $H$ when rendering at a specific checking viewpoint $P_C$ in the image $\boldsymbol{\Omega}(P_C)$. 
It is crucial to prevent the emergence of false positives of hidden watermarkers in the renderings of Gaussians, generated at checking viewpoint by the original generative network $F_{\boldsymbol{\Theta}_0}$, as $\boldsymbol{\Omega}_{0}(P_C)$, which lacks steganographic features. Even if the images rendered by $\boldsymbol{\Omega}_{0}$ and $\boldsymbol{\Omega}$ visually appear identical, we aim to minimize the following contrastive loss term:
\begin{equation}
% \mathcal{L}^+_{{dec}} = | F_{\psi}({\boldsymbol{\Theta}}(P)) - I |, \quad
% \mathcal{L}^-_{{dec}} =  | F_{\psi}({\boldsymbol{\Theta}}_{0}(P)) - \varnothing |,
\mathcal{L}_{{dec}^+} = | F_{\boldsymbol{\Psi}}(\boldsymbol{\Omega}(P_C)) - H |, \quad
\mathcal{L}_{{dec}^-} =  | F_{\boldsymbol{\Psi}}(\boldsymbol{\Omega}_0(P_C)) - \varnothing |,
\label{EQ:DEC1}
\end{equation}
where $\varnothing$ is a meaningless constant image that can be pre-defined by the users.
Effectively, $L_{{dec}^+}$ serves as a regularization term, aiding the decoder in recovering the embedded image patterns based on the rendering obtained from the model. On the contrary, $L_{{dec}^-}$ prevents the decoder from erroneously generating any seemingly reasonable image patterns when rendering is given from the standard generated Gaussians without any hidden signal. The decoder $\boldsymbol{\Psi}$ can be conveniently implemented as U-Net to decode $H$ into the form of a 2D image.
Although the above discussion is primarily focused on hiding images, our framework can easily be extended to embed other modalities such as strings, text, and even audio, all of which can be represented as 1D vectors. We can simply modify the architecture of $\boldsymbol{\Psi}$ to have a 1D prediction branch.

\subsection{Preserving Perceptual Identity}
\label{sec:train}
\textbf{Overall Loss.}
We retain the standard photometric error in steganography learning to maintain the Gaussian rendering fidelity across any views between the steganographic one and the original one:
% $\mathcal{L}_{rgb} = | {\boldsymbol{\Theta}}(P) - {\boldsymbol{\Theta}}_{0}(P) |$.
% Efficient Tuning
    % [Render Loss]
$\mathcal{L}_{rgb} = | \boldsymbol{\Omega}(P) - \boldsymbol{\Omega}_0(P) |$.
The overall training loss of the framework can be formulated as follows, given the input reference image $I$ and hidden image $H$:
\begin{equation}
\mathcal{L}(I, H; \boldsymbol{\Theta}, \boldsymbol{\Phi}, \boldsymbol{\Psi} ) =  \lambda_1\mathcal{L}_{{dec}^+} + \lambda_2\mathcal{L}_{{dec}^-} +  \lambda_3 \mathcal{L}_{rgb}.
\label{EQ:OVERALL}
\end{equation}
where $\lambda_1, \lambda_2, \lambda_3$ is the trade-off coefficients. The parameters that are optimized by the above loss is listed as the generative model $\boldsymbol{\Theta}$, the hidden embedder $\boldsymbol{\Phi}$ and hidden decoder $\boldsymbol{\Psi}$.
Please note that while the parameters of the Gaussian $\boldsymbol{\Omega}$ are present in the loss formula, during the optimization of these parameters, we directly update the generative network $\boldsymbol{\Theta}$, thereby indirectly influencing the generated Gaussian by $\boldsymbol{\Omega}=F_{\boldsymbol{\Theta}} (I)$,

\textbf{Adaptive Gradient Harmonisation.}
Given our objective to embed information without altering the visual perception of the rendered output, one might intuitively consider penalizing the deviation between $ \boldsymbol{\Theta}$ and $\boldsymbol{\Theta}_{0}$ as an effective regularization. However, we have found that naively incorporating penalties for deviations of all weights impedes the generative network's ability to alter its weights for steganographic purposes. Instead, we take into account two key insights: 1) The quality of the rendering and the hidden information both exert influence on the generative network, leading to potentially conflicting demands as the network must conceal hints about the hidden information in certain aspects of the rendering. 2) The weights of the generative network do not equally contribute to the quality of the GS rendering and exhibit strong sparsity. Inspired by these insights, in what follows, we introduce an adaptive gradient harmonization strategy to embed hidden information into specific weight groups of the generative network, where the gradient update requirements for rendering and hidden information are aligned.

% \begin{wrapfigure}{r}{0.48\textwidth}
%     \vspace{-0.4cm}
%     \centering
%      \includegraphics[width=0.9\linewidth]{imgs/gradient.pdf}
%   \caption{
%   \textcolor{red}{TODO}
%   {Illustration of Ambiguous Task in Practice.}
%   }
%  \vspace{-0.4cm}
% \label{fig:robust}
% \end{wrapfigure}

Formally, given the weights ${\boldsymbol{\Theta}}\in\mathbb{R}^N$, we compute a gradient mask ${\mathcal{M}}\in\mathbb{R}^N$ that indicates the whether the gradient w.r.t. rendering and hidden embedding is harmony for the weights:
%.
% Specifically, for each weight
% bu${w}_i \in {{\boldsymbol{\Theta}_{0}}}$, we compute:
\begin{equation}
% [TO MODIFY]
% {m}_i =  \frac{{|{w}_i}|^{-\alpha}}{\sum_{i}^{N} |{w}_{i}|^{-\alpha}},
% {m}_i =  \frac{{|{\boldsymbol{\Theta}_i}}|^{-\alpha}}{\sum_{i}^{N} |{w}_{i}|^{-\alpha}},
{\mathcal{M}} = \mathbb{I}\big(\operatorname{Cos}(
\frac{\partial \mathcal{L}_{rgb}}{\partial \boldsymbol{\Theta}}, 
\frac{\partial (\mathcal{L}_{{dec}^+} + \mathcal{L}_{{dec}^-})}{\partial \boldsymbol{\Theta}}
) > 0 \big)
\label{EQ:REG1}
\end{equation}
% where the magnitude $|\cdot|$ of
% each model weight $w_i$ serves as its individual importance indicator, and $\alpha>0$ is a scaling factor of power controlling the
% relative distribution of importance across the weights.
where $\operatorname{Cos(\cdot, \cdot)}$ denotes the cosine similarity between the two gradient components, $\mathbb{I}$ is an indicator function that is equal to 1 if the condition is true and 0 otherwise. 
We mask the gradient as $\frac{\partial \mathcal{L}}{{\partial \boldsymbol{\Theta} }} \odot {{\mathcal{M}}}$ when optimizing $\boldsymbol{\Theta}$ based on the total loss $\mathcal{L}$, where $\odot$ is a Hadamard product.
Effectively, the gradients related to information embedding, which are inconsistent with the objective of maintaining the rendering from the pre-trained generative model that tends to generate high-quality GS representations, are "masked out" on those conflicting weights. This is done to minimize the impact of steganographic learning on the visual quality of the rendering.
% more significant weights are ``masked out'' to minimize the impact of steganographic learning on the rendered visual quality.
% Note that identifying the weight saliency by its amplitude is also a common strategy in model pruning techniques, but here we hold one more insight into its significance to the ultimate rendering views. 

\section{Experiment}
\subsection{Experimental Setup}

\textbf{Dataset}
We train our model using a filtered subset of the Objaverse dataset~\cite{deitke2023objaverse111}. This subset excludes many low-quality 3D models (for instance, partial scans and those lacking textures) that are present in the original Objaverse dataset. This process results in a final collection of approximately 80K 3D objects. 
 We randomly select 100 objects from the dataset for training, while a separate set of 100 objects, unseen during the training process, is reserved for testing.
We render RGBA images from 40 camera views at a resolution of $256 \times 256$ for both training and evaluation.
For the hidden images, we construct a dataset composed of watermarked images, featuring 100 emoji images. This dataset incorporates 100 routinely utilized image.

\textbf{Evaluation Metrics.} 
The quality of the hidden information that the decoder recovers, including the Peak Signal-to-Noise Ratio (PSNR) and Structural Similarity (SSIM), is evaluated. For the evaluation of the final  rendered images by GS, PSNR, SSIM, and LPIPS are employed. All metrics are calculated on the test set and averaged across all scenarios and embedded images~\cite{chen2022aug,li2022knowledge,liang2021unsupervised}.
% All 40 views are used to evaluate the rendering results.
For subjective evaluation, 360-degree rotational videos of 3D Gaussians generated by different methods are rendered for a collection of 30 images. 30 samples from a mix of random methods are presented to each volunteer and they are asked to score based on the overall visual quality of the image. The results are collected from 20 volunteers.

\textbf{Implementation.}
Our approach is deployed over an advanced image-to-3DGS model, LGM~\cite{tang2024lgm}, which serves as the Gaussian generator. 
% Considering the cost of fine-tuning the base model, we employ LoRA~\cite{} for efficient fine-tuning.
A simple U-Net is utilized as the decoder for hidden information.
Upon the Gaussian generation foundation, we fine-tune a LoRA~\cite{hu2021lora} for each watermark. This training routine typically necessitates training over N=100 objects for approximately 30 epochs, which takes around 20 minutes, and subsequently, it can be generalized to other unseen objects.
We employ the AdamW optimizer for optimization, with a learning rate of $1e-4$.
 For hyper-parameters in Eq.~(\ref{EQ:OVERALL}), we set the weight $\lambda_1=0.3, \lambda_2=1, \lambda_3=0.1$ for all experiments.

\begin{figure*}[t]
  \centering
  \includegraphics[width=0.95\linewidth]{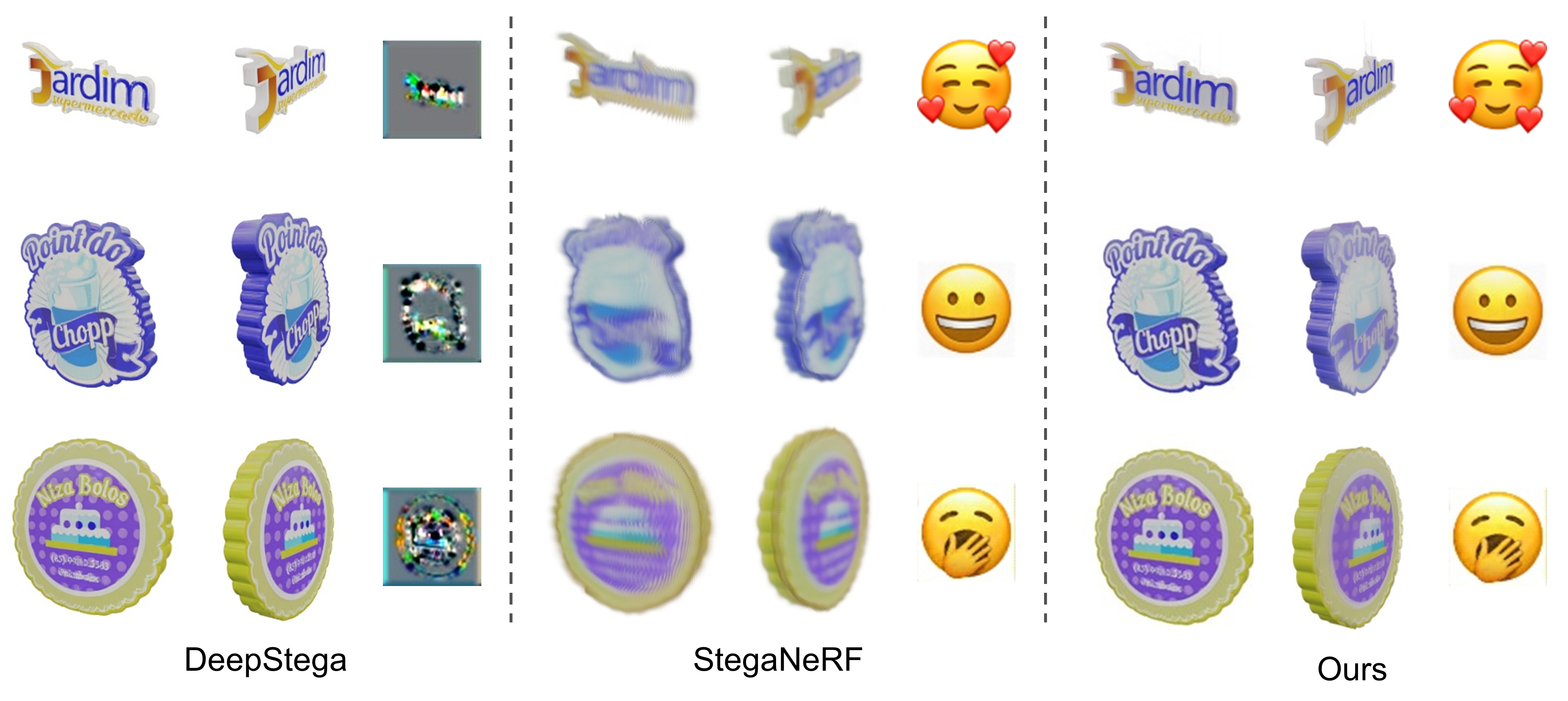}
  \caption{
Qualitative comparison on the test objects of the \obj~dataset.
Within each column, we show the rendering images on check pose and and the recovered hidden images.
  % Qualitative comparison of 
  %  Examples of three available ground-truth expert labels and sampled segmentation masks are provided..
  }
  \label{fig:exp1}
\end{figure*}

\begin{table}[htbp]
  \centering
  \caption{
   Quantitative comparison in rendering and hidden information recovery.
  % {\it Standard NeRF} is the initial NeRF ${\boldsymbol{\Theta}_{0}}$ with standard training (upper-bound rendering performance).
  %$\uparrow$/$\downarrow$ means that larger/smaller numbers denote better performance.
  %
  % Prior 2D steganography fails after NeRF training while \model~successfully embeds and recovers hidden information with minimal impact on the rendering quality.
  % Results are averaged over the selected LLFF and NeRF-Synthetic scenes.
  }
   \resizebox{0.95\linewidth}{!}{
    \begin{tabular}{l|cccc|cccc}
    \toprule
    \multirow{2}[4]{*}{Method} & \multicolumn{4}{c|}{Rendering}                   & \multicolumn{4}{c}{Hidden Recovery} \\
\cmidrule{2-5} \cmidrule{6-9}      \multicolumn{1}{c|}{} & \multicolumn{1}{c}{PSNR\,$\uparrow$} & \multicolumn{1}{c}{SSIM\,$\uparrow$} & \multicolumn{1}{c}{LPIPS\,$\downarrow$} & \multicolumn{1}{c|}{Subj.\,$\uparrow$} & \multicolumn{1}{c}{PSNR\,$\uparrow$} & \multicolumn{1}{c}{SSIM\,$\uparrow$} & \multicolumn{1}{c}{LPIPS\,$\downarrow$} & \multicolumn{1}{c}{Subj.\,$\uparrow$} \\
    \midrule
    Init. Render     &  20.48      & 0.8522        &  0.1181   &   5.00           &    N/A          &   N/A             &      N/A          &N/A   \\
    \midrule
 LSB~\cite{chang2003finding}          & \cellcolor{top2}20.45      & \cellcolor{top2}0.8518      & \cellcolor{top1}0.1185     &    \cellcolor{top2}3.95          & 8.36         & 0.2091     & 0.5379       & 1.20 \\
    DeepStega~\cite{baluja2017hiding}    & \cellcolor{top3}20.43       & \cellcolor{top3}0.8513     & \cellcolor{top3}0.1197    &   \cellcolor{top3}3.21           & \cellcolor{top3}12.11       & \cellcolor{top3}0.2847   & \cellcolor{top3}0.4432       & \cellcolor{top3}1.83  \\
    StegaNeRF~\cite{li2023steganerf}  & 18.53       &     0.8362         &  0.1756            &  2.30            & \cellcolor{top2}31.87        & \cellcolor{top2}0.9659    & \cellcolor{top2}0.0114       &  \cellcolor{top2}3.25\\
 \model~(Ours)         & \cellcolor{top1}20.45 & \cellcolor{top1}0.8519 & \cellcolor{top2}0.1189 & \cellcolor{top1}4.11 &\cellcolor{top1}32.97 & \cellcolor{top1}0.9808& \cellcolor{top1}0.0082 & \cellcolor{top1}3.67 \\
    \bottomrule
    \end{tabular}%
    }
  \label{tab:exp1}%
\end{table}%

\subsection{Embedding 2D Visual Contents as Hidden Information}
% \paragraph{Watermark Images}

% We collected a 100-image data set, which contains 100 unique 64×64 emoji expressions, such as happy, angry, etc. Image for this dataset are collected by the authors
% We train our personalized text to 3D models on the image collections released by the authors of [38]. This dataset consists of 30 different image collections with 4-6 casual captures of a wide variety of subjects (dogs, toys, backpack, sunglasses, cartoon etc.). We additionally capture few images of some rare objects (like "owl showpiece" in Fig. 4) to analyze performance on rare objects. Further, we optimize each 3D model on 3-6 prompts to demonstrate 3D contextualizations.

\textbf{Baseline Models.}
Due to the lack of prior study on steganography specifically for Gaussian Splatting, we consider a baseline from 2D image steganography by fine-tuning large generation models from scratch with watermarked images. We implement two off-the-shelf steganography methods including a traditional machine learning approach, Least Significant Bit (LSB~\cite{chang2003finding}), and a deep learning pipeline, DeepStega~\cite{baluja2017hiding}. We make enhancements to ongoing work, StegaNeRF~\cite{li2023steganerf}, that is dedicated to embedding implicit yet recoverable image watermarks into NeRF, by adapting it to Gaussians. These baseline models serve as crucial benchmarks for evaluating the effectiveness of our proposed approach in the context of Gaussian Splatting steganography. Additionally, the adaptation of existing techniques to this novel domain provides valuable insights into the unique challenges and opportunities presented by Gaussian Splatting in information hiding tasks.

Tab.\ref{tab:exp1} contains quantitative results tested on the selected 100 objects in the Objaverse dataset. Despite the difficulty in recovering embedded information through 2D steganography methods and maintaining the quality of rendering through StegaNeRF, our model minimizes the impact on the rendering quality as measured by PSNR. Fig.\ref{fig:exp1} provides a comparison of steganography on three objects in the \obj~dataset using our model. In comparison to other techniques, the renderings produced by our proposed model retain the details of the original rendering unaffected by steganography to the greatest extent, while accomplishing precise recovery of the watermark. These results underscore the superiority of our approach in balancing the trade-off between information hiding capacity and visual fidelity, a crucial aspect in practical applications of 3D steganography.

\begin{figure*}[t]
  \centering
  \includegraphics[width=0.95\linewidth]{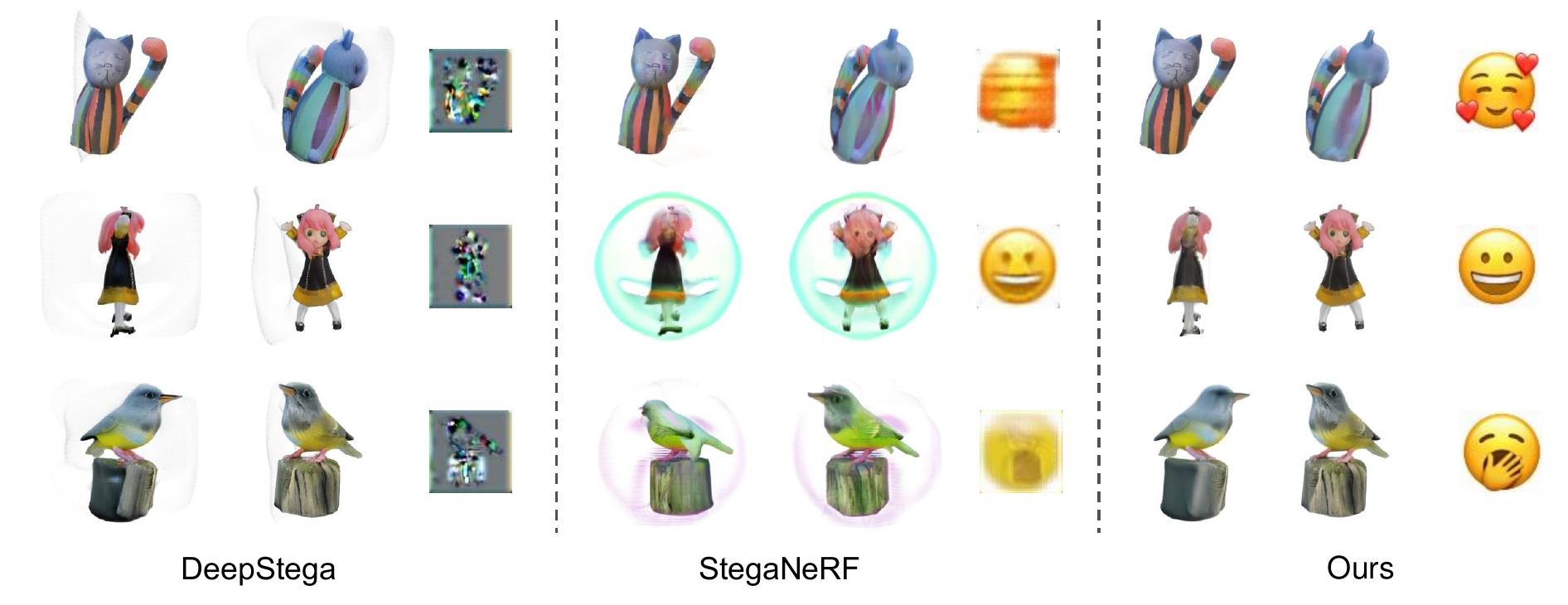}
  \caption{
Quantitative comparison on widely-used test images by image-to-3D models.
Within each column, we show the rendering and the recovered hidden images.
  }
  \label{fig:exp2}
\end{figure*}

Our framework undergoes rigorous testing within real-world image-to-3D deployment scenarios. As depicted in Fig.\ref{fig:exp2}, we utilize our method to embed watermarks in a variety of prevalent use-cases and contrast its effectiveness with other leading-edge methods to study the generalizability\cite{ye2023featurenerf,wang2022attention,li2022domain,li2021consistent} of \model. In comparison to alternative techniques, the renderings generated by our proposed \model~optimally preserve the details present in the original, unwatermarked rendering while successfully extracting the watermark for even unseen objects\cite{wang2022attention,li2021unsupervised,zhang2021generator,xu2022afsc,ding2022unsupervised}. This demonstrates the applicability and value of our method in practical 3D asset production environments. Furthermore, the ability of our model to generalize to unseen objects highlights its potential for widespread adoption in various industries, from entertainment to design, where protecting intellectual property in 3D assets is of paramount importance.

% \paragraph{Can \model~generalize to unseen objects?}

% \paragraph{Can \model~applicable in the wild?}

\subsection{Embedding Multimodal Contents as Hidden Information}
\begin{figure*}[t]
  \centering
  \includegraphics[width=0.92\linewidth]{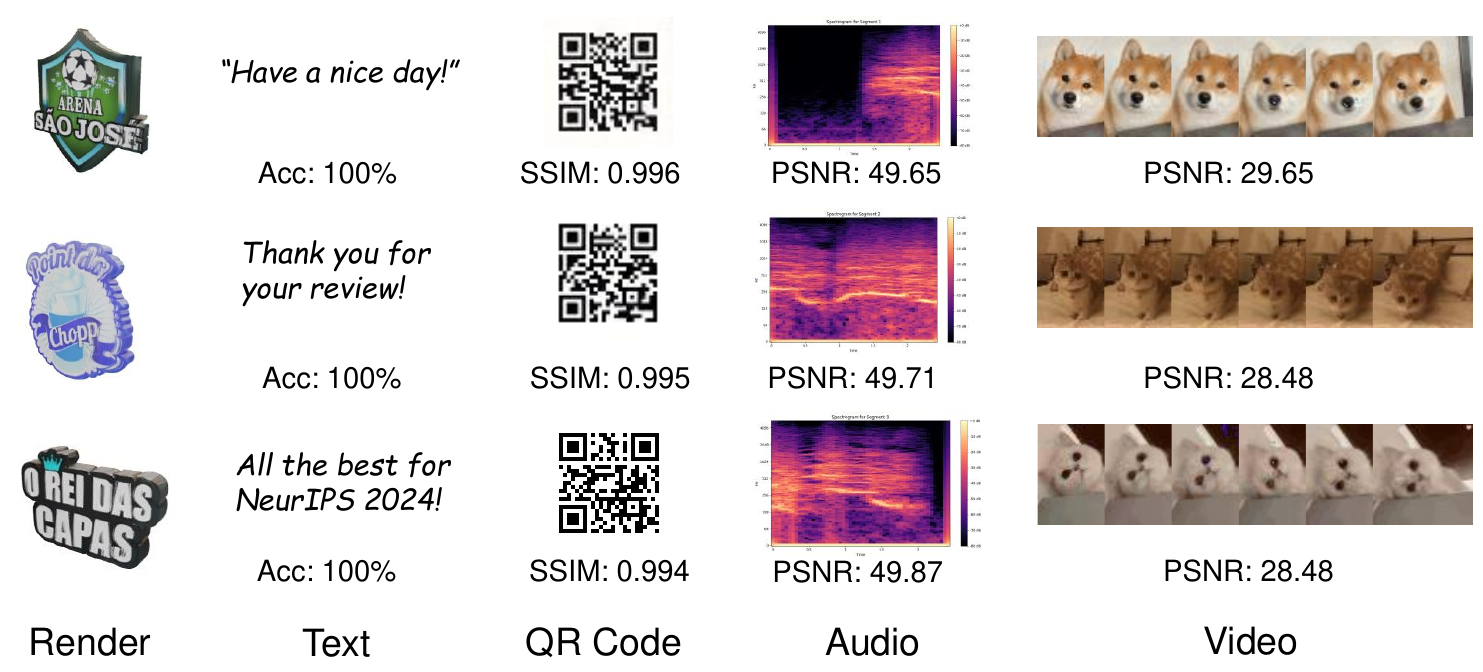}
  \caption{
 Quantitative results of \model~with multimodal information being embedded.
  }
  \label{fig:exp3}
\end{figure*}

We further explore the capabilities of \model\ in integrating multimodal hidden information such as text, QR codes, audio, and video into generated objects. Compared to traditional image watermarking techniques, utilizing multimodal data provides a richer and more comprehensive information set. However, it also introduces the challenge of managing an increased volume of embedded information. To address this, we enhance the decoder network by implementing a modality-specific decoder for each type of input modality. This allows for more effective recovery and utilization of the embedded information.
Figure~\ref{fig:exp3} illustrates the successful recovery of multimodal embedded signals from three generated objects, demonstrating the effectiveness of our approach. We use different metrics to evaluate the quality of recovery for different modalities: ACC (Accuracy) for text embedding recovery, SSIM (Structural Similarity Index) for QR code watermark recovery, and PSNR (Peak Signal-to-Noise Ratio) for audio recovery. Our results indicate that these metrics are suitable for assessing the quality of recovered information across various modalities.

Furthermore, we experiment with embedding a 16-frame video into assets represented by Gaussian representations. The success of this experiment suggests that the \model\ framework can be easily adapted to accommodate multimodal information with high recovery performance, all while maintaining the rendering quality of the generated objects. This flexibility and efficiency in handling multimodal information open up numerous potential applications for watermarking in generative 3D Gaussian contexts. For instance, this could be particularly useful in digital rights management, where multimodal watermarks could provide additional layers of security and authentication. Additionally, in fields like entertainment or education, multimodal embedding could enable the creation of interactive and enriched content by seamlessly integrating various types of media into a single 3D object.

\begin{figure*}[t]
\centering
    \makeatletter\def\@captype{table}\makeatother
    \caption{Ablation study on the proposed key components.}
    \setlength{\tabcolsep}{4.2mm}
	% \resizebox{0.95\linewidth}{!}
 {
    \begin{tabular}{l|cc|cc}
    \toprule
    \multirow{2}[4]{*}{Method} & \multicolumn{2}{c|}{Rendering} & \multicolumn{2}{c}{Hidden Recovery} \\
\cmidrule{2-3} \cmidrule{4-5}     \multicolumn{1}{l|}{} & \multicolumn{1}{c}{PSNR\,$\uparrow$} & \multicolumn{1}{c|}{SSIM\,$\uparrow$} & \multicolumn{1}{c}{PSNR\,$\uparrow$} & \multicolumn{1}{c}{SSIM\,$\uparrow$}\\
    \midrule
   No All Components      & 18.39        & 0.7904       & 27.32        & 0.8870  \\
    No DINOv2      & \cellcolor{top3}20.04        & \cellcolor{top3}0.8507      & \cellcolor{top3}30.60        & \cellcolor{top3}0.9695  \\
 No Cross Attention & 19.66        &0.8395       & 30.15        & 0.9669  \\
No Gradient Harmoni. & \cellcolor{top2}20.11        & \cellcolor{top2}0.8529       & \cellcolor{top2}31.17        & \cellcolor{top2}0.9629 \\
    Full Model (Ours)        & \cellcolor{top1}20.45        & \cellcolor{top1}0.8519       & \cellcolor{top1}32.97        & \cellcolor{top1}0.9808  \\
    \bottomrule
    \end{tabular}%
   % \vspace{-2cm}
 
 }
 \label{tab:abl}

\end{figure*}

\subsection{Ablation Studies}
% The effect of removing each component of \model~is presented in Tab.~\ref{tab:abl}.
% The variant \textit{No All Components} discards all the proposed additions and only retains the standard rendering and steganographic loss. \textit{No DINOv2} eliminates the introduced DINOv2 as the feature extractor for the hidden image and instead employs CLIP Image encoder. \textit{No Cross Attention} and \textit{No Gradient Harmoni.} respectively eliminate the cross-attention of hidden information on the intermediate features of Gaussian generation and the introduced gradient harmonization strategy.
% It appears that when any component is removed, the performance drops accordingly, revealing the effectiveness of our design.
In Tab.~\ref{tab:abl}, we present an analysis of the effect of removing each component of \model. This analysis is crucial in understanding the contribution of each component to the overall performance of the system.
The variant termed as \textit{No All Components} discards all the proposed additions to the system. This version only retains the fundamental rendering and steganographic loss, providing a baseline for comparison.
On the other hand, the variant \textit{No DINOv2} eliminates the newly introduced DINOv2. This component is used as the feature extractor for the hidden image. Instead of DINOv2, this variant employs the CLIP Image encoder, providing a different approach to feature extraction.
The variants \textit{No Cross Attention} and \textit{No Gradient Harmoni.} respectively eliminate the cross-attention of hidden information on the intermediate features of Gaussian generation and the newly introduced gradient harmonization strategy. These variants help us understand the impact of these specific features on the system.
Through this analysis, it becomes clear that when any component is removed, the performance drops accordingly, underscoring the effectiveness of our design in each component.

\subsection{Discussion}

\begin{figure}[!t]  
  \centering
  \subfloat[Varying JPEG compression rate] %第二张子图
  {
      \begin{minipage}[t]{0.4\linewidth}
          \centering      %子图居中
          \includegraphics[width=\linewidth]{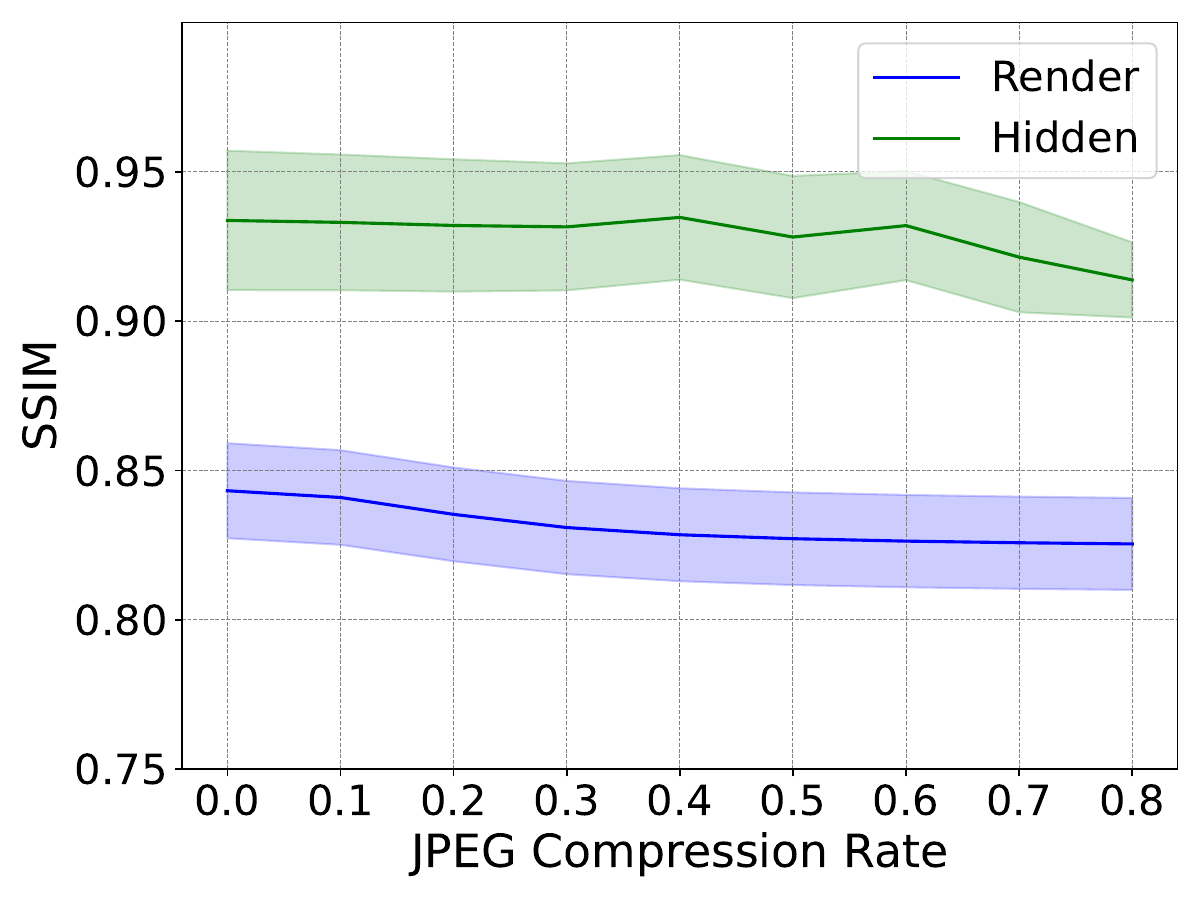} 
          %\vspace{-1em}
      \end{minipage}
  }%
   \subfloat[Varying Gaussian blur std.] 
  {
      \begin{minipage}[t]{0.4\linewidth}
          \centering          %子图居中
          \includegraphics[width=\linewidth]{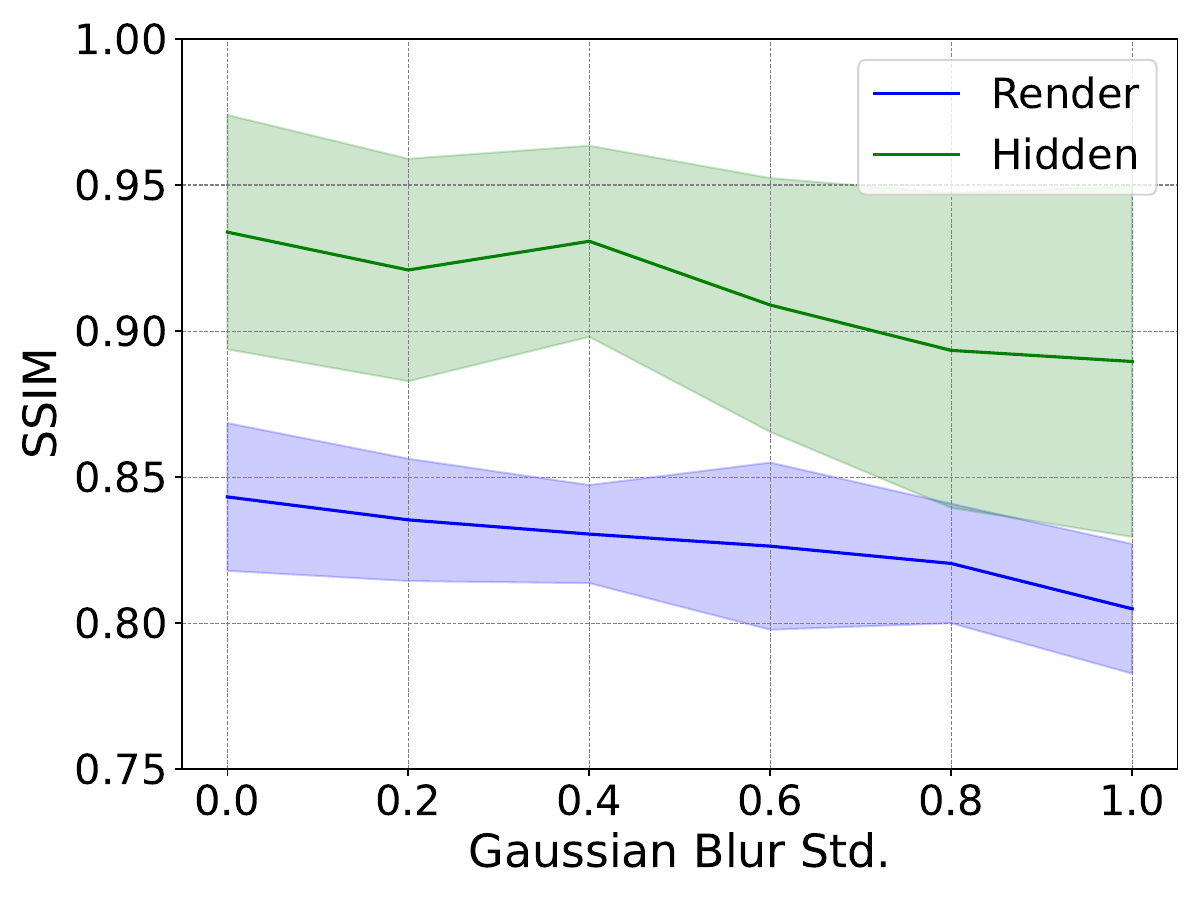} 
           %\vspace{-1em}
      \end{minipage}%
  }
%   %\vspace{-0.5em}
  \caption{
  Analysis of robustness over (a) JPEG compression and (b) Gaussian blur. 
  We provide the SSIM of rendered views (blue) and recovered hidden images (green).
  } 
%   %\vspace{-1.5em}
\label{fig:robust}  
\end{figure}

\textbf{How about Robustness?}
As shown in Fig.\ref{fig:robust}, we can observe that \model~is robust against common perturbations, such as JPEG compression and Gaussian noise.  The curves are representative of mean accuracies that have been computed across selected scenes. The shaded regions, on the other hand, signify a range of 0.5 standard deviation. This suggests a certain degree of variability within the data even with the limited data quality~\cite{pan2023learning,sun2022few,zhang2021generator}. The results of this analysis strongly suggest that the ability of the \model~to recover hidden information remains consistent and resilient, even when exposed to a wide variety of JPEG compression levels and deteriorations caused by Gaussian blur. This  robustness extends the practical applicability of \model~to real-world scenarios where image manipulations and transformations are common, ensuring the integrity of embedded information across various digital environments and transmission channels.

% We will include it in the revision.

% \begin{figure}[!h]  
%     \centering
    
%     \begin{subfigure}[h]{0.48\linewidth}
%         \centering
%         \includegraphics[width=\linewidth]{Rebuttal/curves_1.pdf}
%         % \caption{Render\_PSNR}
%         % \label{fig:render_psnr}
%     \end{subfigure}
%     \hfill
%     \begin{subfigure}[h]{0.48\linewidth}
%         \centering
%         \includegraphics[width=\linewidth]{Rebuttal/curves_2.pdf}
%         % \caption{Hidden\_SSIM}
%         % \label{fig:hidden_ssim}
%     \end{subfigure}
%     \vspace{-0.5em}
%     \caption{Ablation on the resolutions of embedded images by the average results over the selected scenes from the LLFF dataset.}
%     \vspace{-1em}
%     \label{fig1}
% \end{figure}

\textbf{Limitations.} 
The current implementation is limited by the computational efficiency of large generative Gaussian model, since the steganographic training stage requires performing gradient descent updates through the large reconstruction model.
Another limitation is the recovery accuracy of the injected hidden information varies across different 3D scenes.
More systematic analysis is required to understand the underlying factors that make some scenes easier to hide information than others.
% The current implementation of our approach encounters certain limitations, the most noteworthy of which pertains to the computational efficiency of the large generative Gaussian model. As the steganographic training stage necessitates performing gradient descent updates via the extensive reconstruction model, the process becomes computationally intensive, thereby restricting prompt results.
% Another restriction of our implementation is the inconsistent recovery accuracy of the injected hidden information across different 3D scenes. This inconsistency can be attributed to the varying characteristics of different scenes, which might influence the efficacy of the steganographic process.
% These limitations indicate that a more systematic and comprehensive analysis is required to understand the underlying factors that render some scenes more conducive to hiding information than others. By investigating these factors, we can enhance the algorithm's efficiency and accuracy, thereby improving its overall performance.

\textbf{Broader Impacts.} 
Propagating 3D GS representations requires tracing technology, a challenge that this paper is poised to address. The tracking method is instrumental in fields such as image rendering and digital watermarking. It enhances our comprehension of complex or unclear content embedded in 3D representations. Proper tracking can improve accuracy in uncovering hidden information and decision-making in digital media manipulation, facilitating precise data retrieval and manipulation planning. Additionally, ensuring the safety of this propagation process is a cornerstone for data privacy and security, thereby fostering trust and user satisfaction.

\section{Conclusion}
This study pioneers the exploration of elevating steganographic information embedding originally designed for image generation, to the emerging 3D content generation like Gaussian Splatting. Our proposed model, named \model, subtly embeds copyright or proprietary information into the rendering of generated 3D assets while preserving the visual quality of the rendering. This innovative method has been rigorously evaluated across various potential deployment scenarios, including embedding image watermarks as well as multimodal watermarks such as text, QR codes, and videos for objects both within and outside the benchmark dataset. This process provides valuable insights and establishes a clear direction for future research endeavors. Essentially, this approach signifies significant progress in our efforts to address the emerging challenge of seamlessly integrating customizable, almost imperceptible, and recoverable information into generated 3D Gaussians.

\newpage

% \section*{References}
% \medskip
\small{
\normalem
\bibliographystyle{abbrv}
\bibliography{reference}
}

\newpage
\end{document}